\documentclass{article}

\usepackage{amsmath}
\usepackage{amsthm}
\usepackage{mathtools}
\usepackage{xargs}
\usepackage{subfig}
\usepackage{authblk}

\usepackage[margin={2cm,2cm}]{geometry}
\usepackage{multicol}
\usepackage{natbib}
\usepackage[utf8]{inputenc} 
\usepackage[T1]{fontenc}    
\usepackage{hyperref}       
\usepackage{url}            
\usepackage{booktabs}       
\usepackage{amsfonts}       
\usepackage{nicefrac}       
\usepackage{microtype}      

\newcommand{\prs}[2]{\ensuremath{\phi_{#1,#2}}}
\newcommand{\prva}{\ensuremath{\phi_\alpha}}

\newcommand{\mfrac}[3]{\ensuremath{\frac{m_{#1,#2,#3}}{M}}}
\newcommand{\setalphaaccuracies}[1]{\ensuremath{\{\phi_{\alpha,i}\}_{i=1}^#1}}
\newcommand{\setbetaaccuracies}[1]{\ensuremath{\{\phi_{\beta,i}\}_{i=1}^#1}}
\newcommand{\decisioneventcounts}{\ensuremath{\{m_{\ell_1,\ell_2,\ell_3}\}\,}}
\newcommand{\phivars}{\ensuremath{\{\phi_{\square}\}}}
\newcommand{\varscorrtwo}{\ensuremath{\{\Gamma_{i,j;\ell}\}}}
\newcommand{\varscorrthree}{\ensuremath{\{\Gamma_{i,j,k;\ell}\}}}
\newcommand{\varscorrn}{\ensuremath{\{\Gamma_{i,j,k,\ldots,n;\ell}\}}}
\newcommand{\acoeff}{\ensuremath{a(\decisioneventcounts)\,}}
\newcommand{\bcoeff}{\ensuremath{b(\decisioneventcounts)\,}}
\newcommand{\ccoeff}{\ensuremath{c(\decisioneventcounts)\,}}
\newcommand{\nlabeltrue}{\ensuremath{n(\ell_{\text{true}})}}
\newcommand{\labeltrue}{\ensuremath{\ell_{\text{true}}}}
\newcommand{\elltrue}{\ensuremath{\ell_{\text{true}}}}

\newtheorem{theorem}{Theorem}

\newtheorem{conjecture}{Conjecture}

\begin{document}
\title{Independence Tests Without Ground Truth for Noisy Learners}
\author[1]{Andr\'es Corrada-Emmanuel}
\author[2]{Edward Pantridge}
\author[2]{Edward  Zahrebelski}
\author[2]{Aditya Chaganti} 
\author[2]{Simeon Simeonov}
\affil[1]{Data Engines}
\affil[2]{IPM.ai}
\date{}
\maketitle
\date{}

\begin{abstract}
Exact ground truth invariant polynomial systems can be written
for arbitrarily correlated binary classifiers. Their solutions give estimates
for sample statistics that require knowledge of the ground truth of the correct
labels in the sample. Of these polynomial systems, only a few have been solved in closed form. 
Here we discuss the exact solution for independent binary classifiers - resolving an outstanding problem that
has been presented at this conference and others. Its practical applicability is
hampered by its sole remaining assumption - the classifiers need to be independent in their 
sample errors. We discuss how to use the closed form solution to create a 
self-consistent test that can validate the independence assumption itself 
absent the correct labels ground truth. It can be cast as an algebraic geometry conjecture for binary
classifiers that remains unsolved. A similar conjecture for the ground truth invariant algebraic
system for scalar regressors is solvable, and we present
the solution here. We also discuss experiments on the Penn ML Benchmark classification tasks that provide
further evidence that the conjecture may be true for the polynomial system of binary classifiers.
\end{abstract}

\section{Introduction}

Self-assessment problems are everywhere in science and technology. For AI algorithms the
topic appears with various names. This conference has the subject heading "Unsupervised Ensemble
Learning". In this paper we will call it \emph{ground truth inference} (GTI) to highlight
the evaluation or monitoring aspect of the task - infer sample statistics that require
knowledge of the correct answers given by an ensemble of algorithms. Since this methodology
is non-parametric in the inference or recovery process, it cannot generalize its findings
but merely evaluate or monitor specific samples.

The problem of GTI for independent binary classifiers was first discussed by \citet{Dawid79}. Their
work further motivated \citet{Raykar2010} to develop a Bayesian approach to the problem.
Spectral methods have also been developed and, in this conference, the work of \citet{Jaffe2015}
gave a partial solution to the independent binary classifiers problem. This paper will discuss 
an exact, closed solution obtained by an algebraic approach (Theorem \ref{th:independent-solution}) - 
construct ground-truth invariant polynomial systems relating the unknown ground truth statistics - 
the sample prevalence of one label \prva, and the sample label accuracies (\setalphaaccuracies{3},
and \setbetaaccuracies{3}) to the label decision event counts,
\decisioneventcounts. Given how many times the classifiers voted with one of the eight possible
ways they could carry out binary classification, the \decisioneventcounts, we obtain, in a fully
non-parametric fashion, two algebraic point solutions for the unknown ground truth statistics with 
one of them guaranteed to be the true one.

We will also discuss and use another polynomial system - the exact polynomial description of the 
\decisioneventcounts in terms of the prevalence and the label accuracies of the classifiers
that we have mentioned but also the decision error correlations on the sample (Theorem \ref{th:general-system}). 
No one can solve this larger system, but it can be used to develop detection tests that can verify if,
in fact, the classifiers are independent in their errors.

The non-parametric character of these polynomial systems has strengths and weaknesses. Unlike the Bayesian
approach to GTI, for example, it makes no assumptions about the functional form needed
to explain the sample errors. This expands the practical application of GTI algorithms by 
eliminating theoretical assumptions  that may bound the applicability of a particular Bayesian formulation,
a deficiency that has been observed in experiments, \citet{Zheng}. And, as we have noted, we obtain an exact 
polynomial formulation of the problem for arbitrarily correlated classifiers (Theorem \ref{th:general-system}).
This allows us to study the case of practical interest - classifiers that are weakly or sparsely correlated.
These are the operating points of well designed engineering systems.
We may not be able to solve this more general formulation, but the ground truth values
that we seek to estimate will be members of the algebraic variety that solves this full polynomial
formulation.  This neatly sidesteps the problem of representation
in AI and allows for strong mathematical conclusions to be made without knowledge of
the underlying ground truth or the processes that could have generated the errors we
now see in a sample. The drawback of this non-parametric approach is clear. 
By focusing on sample statistics alone, we lose any ability to generalize the
sample estimates to other samples

We construct two independence tests for binary classifiers when the ground truth is absent.
The first shows the utility of the algebraic approach by constructing a nearly
perfect detector of non-independence when we have just three classifiers. Applying
Theorem \ref{th:independent-solution} to non-independent classifiers results in an
estimate for the prevalence of either label that contains an irreducible square root.
Since for any finite sample the sample prevalence is always an integer ratio, this can
only happen if the classifiers are non-independent.

The second independence test considers how we could use the
consistency or inconsistency of solutions when Theorem \ref{th:independent-solution} is used on all possible
trios of an ensemble of $n>=4$ classifiers as a way to detect nearly independent classifiers.
We demonstrate, with experiments on a set of binary classification datasets, that the use
of four classifiers and repeated use of Theorem \ref{th:independent-solution} seems to be
related to the label error correlations and that they do not "mix" - poor recovery of the
accuracies for one label that has large correlations is accompanied by good recovery
for the other label where the error correlations are much smaller. This too may have an
algebraic explanation we briefly note later.

\subsection{Ground truth statistics as polynomial systems}

Our first theorem gives the solution to the independent binary classifiers problem.
\begin{theorem}
Given the set of eight integer counts \decisioneventcounts for all
possible ways that three
binary classifiers independent in their errors can output the two labels
$\alpha$ and $\beta$ on a sample they have classified, if $\{m_{\ell_1,\ell_2,\ell_3}>=1\}$, 
and $M=\sum m_{\ell_1,\ell_2,\ell_3},$
the following set of quartic equations has only two point solutions and one of them is the
ground truth prevalence of the $\alpha$ label, \prva, and the classifiers accuracies on each of the
labels, \setalphaaccuracies{3} and \setbetaaccuracies{3},
\begin{gather}
  \mfrac{\alpha}{\alpha}{\alpha}  =  \prva \prs{1}{\alpha} \prs{2}{\alpha} \prs{3}{\alpha} +
    (1 - \prva) (1 - \prs{1}{\beta}) (1 - \prs{2}{\beta}) (1 - \prs{3}{\beta})\\
  \mfrac{\alpha}{\alpha}{\beta}  =  \prva  \prs{1}{\alpha} \prs{2}{\alpha} (1 - \prs{3}{\alpha}) +
    (1 - \prva) (1 - \prs{1}{\beta}) (1 - \prs{2}{\beta}) \prs{3}{\beta}\\
  \mfrac{\alpha}{\beta}{\alpha}  =  \prva  \prs{1}{\alpha}  (1 - \prs{2}{\alpha})  \prs{3}{\alpha} +
    (1 - \prva) (1 - \prs{1}{\beta}) \prs{2}{\beta} (1 - \prs{3}{\beta})\\
  \mfrac{\beta}{\alpha}{\alpha}  =  \prva  (1 - \prs{1}{\alpha}) \prs{2}{\alpha} \prs{3}{\alpha} +
    (1 - \prva) \prs{1}{\beta} (1 - \prs{2}{\beta}) (1 - \prs{3}{\beta})\\
  \mfrac{\beta}{\beta}{\alpha}  =  \prva  (1 - \prs{1}{\alpha}) (1 - \prs{2}{\alpha}) \prs{3}{\alpha} +
    (1 - \prva) \prs{1}{\beta} \prs{2}{\beta} (1 - \prs{3}{\beta})\\
  \mfrac{\beta}{\alpha}{\beta}  =  \prva  (1 - \prs{1}{\alpha})  \, \prs{2}{\alpha} \, (1 - \prs{3}{\alpha}) +
    (1 - \prva) \prs{1}{\beta} (1 - \prs{2}{\beta}) \prs{3}{\beta}\\
  \mfrac{\alpha}{\beta}{\beta}  = \prva  \prs{1}{\alpha}  (1 - \prs{2}{\alpha})  (1 - \prs{3}{\alpha}) +
    (1 - \prva) (1 - \prs{1}{\beta}) \prs{2}{\beta} \prs{3}{\beta}\\
  \mfrac{\beta}{\beta}{\beta}  = \prva  (1 - \prs{1}{\alpha})  (1 - \prs{2}{\alpha}) (1 - \prs{3}{\alpha}) + 
    (1 - \prva) \prs{1}{\beta} \prs{2}{\beta} \prs{3}{\beta}.
\end{gather}
\label{th:independent-solution}
\end{theorem}

The second theorem highlights that sample statistics of the decisions on a sample are complete, and
we can write down the exact polynomial system that can compute the decision event counts in terms
of a finite set of unknown ground truth statistics.
\begin{theorem}
Given the set $\{m_{\ell_1,\ell_2,\ldots,\ell_n}\}$ of decision event counts done by an
ensemble of $n$ binary classifiers that are arbitrarily correlated in their sample errors,
we can write an exact polynomial system that relates the counts to \prva, \setalphaaccuracies{n},
\setbetaaccuracies{n} and the error correlations on the sample, \varscorrtwo \, up to \varscorrn.
\label{th:general-system}
\end{theorem}

The existence of multiple solutions for the polynomial systems, such as the ones in Theorems
\ref{th:independent-solution} and \ref{th:general-system}, follows from their invariance under
ground truth transformations. This is evident for the polynomial system in Theorem \ref{th:independent-solution}
because it is invariant under the following transformations of the ground truth statistics we are solving for,
\begin{align}
    \prva &\longrightarrow 1 - \prva \\
    \{\phi_{\alpha;i} &\longrightarrow \phi_{\beta;i}\}\\
    \{\phi_{\beta;i} &\longrightarrow \phi_{\alpha;i}\}.
\end{align}
So if one set of values for the ground truth statistics solves the polynomial system, 
another set of values - the one obtained by applying the transformations above -
will also solve it.

The practical use of this algebraic approach to GTI is that the polynomial systems
are neither trivial and may have interesting point solutions that greatly simplify the
use of GTI for nearly unsupervised online evaluation of noisy learners as Theorem 1 does. 
In this way, algebraic GTI is no different than the use of error-correcting codes to fix bit flips in modern
computers. All codes have multiple solutions to the decoding problem. Yet they
are widely used in modern computers because their engineering context - well-manufactured
hardware is reliable enough to make a few bit flips much more likely than many - makes them practical
as a way to fix occasional, stray bit flips.

Following the intuition that nearly independent algorithms would produce consistent results
when different trios of them are used to apply Theorem \ref{th:independent-solution}, we
explore here a proposition we call the \emph{four consistent trio solutions conjecture.}
\begin{conjecture}
Given the decisions of four binary classifiers on a sample, if the 4 possible ways to use Theorem
1 on their decision event counts yield consistent estimates for \prva, \setalphaaccuracies{4}, and
\setbetaaccuracies{4} then their 2 and 3 way error correlations on the sample are identically
zero,
\begin{align}
    \{\Gamma_{\alpha,i,j} = & \, 0\} &  \{\Gamma_{\beta,i,j} = & \, 0\}\\
    \{\Gamma_{\alpha,i,j,k} = & \, 0\} &  \{\Gamma_{\beta,i,j,k} = & \, 0\}
\end{align}
\end{conjecture}

\subsection{Organization of the paper}

We finish the introduction placing algebraic GTI in relation to data sketching, compressed sensing,
and algebraic statistics. Section 2 formalizes \emph{ground truth}, \emph{ground truth statistics}, and
\emph{observable statistics} in the context of two GTI tasks - estimating
the accuracy of binary classifiers as we have been discussing and recovering the
precision error covariance matrix for scalar regressors. 
Section 3 constructs an independence test with three classifiers based on
the algebraic nature to the solutions found in Theorem \ref{th:independent-solution}.
This highlights the relevant practical problem: given unknown error correlations, 
can we use the variations in independent model solution to bound them? 
Section 4 tests the four consistent trios conjecture on three Penn ML binary
classification benchmarks. Section 5 partially resolves the conjecture for a similar formulation
on a GTI task for scalar regressors. Finally, a closing section summarizes the work and advocates
for more research into these algebraic methods for measuring AI errors in a non-parametric
fashion.

\subsection{Similarities of algebraic GTI to other fields}

Algebraic GTI has similarities and differences with other areas of research that may be
known by the reader - data sketching, compressed sensing, and algebraic statistics. 

Like data sketching, any specific algebraic GTI polynomial system solves for a specific statistic. 
There is no recovery of the whole ground truth with algebraic GTI. Only statistics of the ground truth
can be recovered. Algebraic GTI is data sketching for AI errors. 
Also, similar to data sketching, the recovery of a ground truth statistic only requires small memory footprints.
Theorem \ref{th:independent-solution} shows how eight integer counters are enough
to recover the sample label accuracies for three independent classifiers. Algebraic GTI
is non-parametric in the same way that data sketching algorithms are non-parametric.

It may seem strange to the reader that statistics of an unknown ground truth could be computed
robustly in the absence of said ground truth, or that we could do it non-parametrically with
minimal theoretical assumptions. But this has been done before. The Good-Turing estimate
is able to estimate the number of unseen species or types in a finite survey by using
the frequency of species observed one time, two times and so on. It then shifts counts in
these observed frequency bins to estimate the count that should be in the zero count bin.
The Good-Turing algorithm can be restated as a series of polynomial equations
that push the observed counts to the zero count bin.

The under-determined nature of ground truth invariant polynomials is also similar to the problems
that Compressed Sensing tries to solve. Indeed, the ground truth invariant linear system for the 
precision error covariance matrix was solved for sparsely correlated scalar regressors 
in \citet{CorradaSchultz2008}
by using $\ell_1$ minimization - a favorite for CS recovery of a signal. 
There is an important difference between CS and algebraic GTI. In GTI you are recovering
the error in the signal, not the signal itself. By focusing on the error of the signal, not the 
signal itself, we can get away with having no knowledge of what produced the signal. 
Algebraic GTI is about error recovery, not signal recovery.

In the 1990s, \citet{Pistone} pioneered the use of algebraic geometry in statistics
by recasting traditional statistical tasks, such as experimental design, as polynomial 
problems. Algebraic GTI also uses the math of algebraic geometry
for a statistical task. The major difference between them is the object of study.
Algebraic statistics has been developed focusing solely on parametric statistics
problems. Algebraic GTI is an example of non-parametric statistics. 
Perhaps future editions of Algebraic Statistics books will include a chapter on this 
non-parametric approach to estimate the statistics of errors made by AI algorithms. 

\section{Terminology and definitions for algebraic GTI}

\subsection{What is ground truth?}

Ground truth in non-parametric estimation of AI errors refers to the correct responses
the noisy learners should have returned for the sample. In the case of binary classifiers
this refers to the true label for each item in the sample that was classified. For regressors,
it refers to the correct scalar, or vector of each item in the sample. The ground truth may
be hidden, unavailable or unknown.

\subsection{What are ground truth statistics?}

\emph{Ground truth statistics} are then statistics of the sample that require knowledge of
the ground truth to be computed. Examples of this for binary classifiers are the true prevalence 
of the  labels in the sample, the accuracies of each classifier on a label, their correlations, 
etc. Examples for regressors are their precision error covariance tensors.

These should be contrasted with \emph{observable statistics} - statistics of the sample
that require no knowledge of the ground truth. Examples of these are the counts with
which the classifiers agreed or disagreed on labels of the sample, and moments of the
predictions for regressors.

Ground truth inference via algebraic methods is based on establishing a polynomial system 
relating the ground truth statistics - to be solved for - with the observable statistics
- readily computed from the sample of noisy decisions. By imposing invariance to
ground truth values, the polynomials eliminate the need for theoretical assumptions
about the process that led to the errors in the sample. No assumptions are
needed besides generic ones like the independence of their errors. Algebraic GTI is trying
to do error recovery, not signal recovery.

\section{Detecting non-zero error correlations}

The algebraic nature of the solution to Theorem \ref{th:independent-solution}
and the full polynomial system of Theorem \ref{th:general-system} allows one to
immediately create a nearly perfect detector for non-independent classifiers on 
the given sample. It is based on the following theorem
\begin{theorem}
Given the decision event counts, \decisioneventcounts, for three binary classifiers,
the independent polynomial system solution for the sample prevalence \prva will
contain the square root of an irreducible polynomial that is not functionally
a perfect square except for the case of zero error correlations.
\label{th:three-independence-test}
\end{theorem}
Because of this theorem, whenever we use the independent polynomial system
to estimate \prva we are almost guaranteed to obtain an irreducible
square root in the solution for \prva. But for any finite size sample the
prevalence is an integer ratio, whether the classifiers are correlated or not in their
sample errors. The appearance of this irreducible square root is almost surely
a sign that the sample does not have all error correlations are zero. We now outline
the algebraic proof for this.

One way to solve a multi-variable polynomial systems is to use Buchberger's algorithm
to obtain an Elimination Ideal, see \citet{Cox}. This Ideal consists of a series of polynomial
equations that have, at its start, a polynomial with a single unknown variable.
Solving this polynomial then allows us to proceed down a chain of other polynomials
to resolve the other unknown variables. This is the constructive algorithm that we use
to prove Theorem \ref{th:independent-solution}. The existence of just two point
solutions for the independent classifiers polynomial system is based on the appearance
of a quadratic for the unknown prevalence \prva at the base of Elimination Ideal.
This quadratic is enormous but it can be represented easily in the following
symbolic form,
\begin{equation}
    \acoeff \prva^2 + \bcoeff \prva +
    \ccoeff = 0,
\end{equation}
where the coefficients \acoeff, \bcoeff, and \ccoeff  are themselves
polynomials in the \decisioneventcounts \linebreak
variables.
So the independent polynomial solution for \prva \, is immediately given by
the quadratic formula and will contain a square root of the form
$\sqrt{b^2 - 4ac}.$

But the ground truth value for \prva \, is an integer
ratio, whether they are correlated or not. So the appearance of a non-reducible 
square root is a clear signal that the independent model cannot be the correct description of
the correlations between the classifiers. But can correlated classifiers
return a solution to the independent model quadratic that is an integer ratio?

Here is where the theoretical advantages of having a full, exact polynomial
description for the observed \decisioneventcounts counts,Theorem \ref{th:general-system},
becomes useful. Start by noting
that the independent polynomial system can be fed back into the \prva quadratic
so we can rewrite \acoeff, \bcoeff, and \ccoeff as polynomials of the unknown
statistics. When we do that the square root term in the quadratic formula
solution becomes,
\begin{multline}
    \sqrt{\bcoeff^2 - 4 \acoeff \ccoeff} = \\
    \sqrt{\left(1-2 \phi _{\alpha }\right){}^2 \left(\phi _{\alpha }-1\right){}^4 \phi _{\alpha }^4 \left(\phi _{1,\alpha }+\phi _{1,\beta
   }-1\right){}^4}
   \sqrt{\left(\phi _{2,\alpha }+\phi _{2,\beta }-1\right){}^4 \left(\phi _{3,\alpha }+\phi _{3,\beta }-1\right){}^4}.
   \label{eq:reducible-independent}
\end{multline}
Since this is the square root of a perfect square, we obtain a self-consistent result - independent classifiers have
integer ratio solutions for the unknown \prva.

If the classifiers are not independent, then we must put in the full polynomial equations assuming non-zero correlations
between the classifiers. Having done so, further algebraic manipulations show that the square root term in the quadratic
formula becomes,
\begin{equation}
    \left( g_1(\phivars) + g_2(\phivars, \varscorrtwo, \varscorrthree) \right)^2
    \left( g_3(\phivars) + g_4(\phivars, \varscorrtwo, \varscorrthree) \right).
\end{equation}
The g polynomials, ($g_1, g_2, g_3, g_4$), are polynomials in the variables of their
arguments. The $g_1$ and $g_4$ terms are reducible, and when the error correlations
are identically zero, this product term reduces to the one in Equation \ref{eq:reducible-independent}.
However, the sum $g_3 + g_4$ is not reducible in general and thus cannot be guaranteed to become
a perfect square. Indeed, to make the general polynomial reducible, you have to assume
all correlations are equal and the classifiers have equivalent accuracies.

Thus the appearance of an irreducible square root term in the solution for the sample
prevalence when we use the independent classifiers polynomial is a nearly perfect detector for the
existence of non-zero correlations. A complete understanding of this result theoretically would then
require enunciating theorems that can pronounce what conditions guarantee, almost surely, no
non-zero correlations can produce a perfect square in the independent model prevalence solution.

This satisfying theoretical result has limited practical use since we expect real samples to
have some correlations, albeit maybe small ones. Must we not use the independent system solutions
even when the correlations are small? What if the solutions have small disagreements?
Hence our interest in Conjecture 1. Can we start by saying that no correlated system would yield
consistent independent solutions? If so, can we then proceed further by using the inevitable small disagreements 
between the trio solutions to understand the size
of the correlations between the classifiers? In this paper we do not fully work out this approach
but we can offer further experimental and theoretical results to motivate the possibility of an
affirmative answer.

\section{Testing the conjecture experimentally}

We now provide experimental evidence for the four consistent trios conjecture
in the case of binary classifiers. We use three of the classification
benchmarks in the Penn ML Benchmarks suite, \citet{Olson2017PMLB}: {\tt twonorm},
{\tt spambase}, and {\tt mushroom}. Our goal is to exhibit cases where the algebraic method 
returns consistent and inconsistent results. We are not asserting the general
applicability of the independent polynomial system solutions, but rather the identification
of settings where it does seem to work reasonably well.

In addition, we will note an interesting phenomenon in the experimental results - using the
consistency of trios seems to detect label correlations somewhat independently. In our
experiments we were able to consistently get good estimates for one label whenever that label
had small error correlations but inconsistent estimates for the other label when it had larger
error correlations. A possible algebraic explanation for this is that the Elimination Ideal
contains two polynomial chains, one for each label, starting at the base quadratic for the
prevalence \prva.

\subsection{Label error correlations}

Label error correlations are also examples of
ground truth statistics. Given the true label for the items
in the sample of size $m$, we can define the n-way error correlation
for a label, $\ell$, as,
\begin{equation}
    \Gamma_{\ell;i,j,\ldots} = \frac{1}{m_{\ell}} \sum_{d=1}^{m} (x_{d,i} - \bar{x_i})(x_{d,j} - \bar{x_j})\ldots
\end{equation}
Here $x_{d,i}$ is an indicator variable that is one if the classifier $i$ correctly labeled item $d$ in
the instances of label $\ell$ in the data. The term $\bar{x_i}$ is the accuracy of the classifier on the $\ell$ items in the sample
and is the definition of the $\phi_{i,\ell}$ ground truth statistic.

The assumption of independence of errors on a given sample is equivalent to saying that all these
correlations are zero for all labels as shown in Conjecture 1.
This is an assertion about the sample correlations not about any underlying process that may be causing 
the ensemble members to agree or disagree on their decisions. 

For ease of illustration, we will be quoting the average, and standard deviations of  2-way error correlations for each
label in our experiments. The experimental verification of the conjecture then consists of
observing that whenever the trio solutions are consistent for all 4 classifiers, the
correlations are small and whenever they disagree, the correlations are large. In other
words, we expect the size of the disagreements to be monotonically related to the correlation
magnitudes. Elucidating how the discrepancy relates to the size of the error correlations
is a subject for future work.

\subsection{Methodology for a single experiment}

We tried our best to create nearly independent classifiers for the three examples
shown in this experimental section. The reader is seeing the experiments 
for which we got closest to that independence condition.
Our approach for inducing as much independence
in the errors as possible was to use these three basic technique: using different
classification algorithms, reducing the overlap in training data, and having
no or little intersection between classifier feature sets.

We trained four classifiers for an experimental run and then apply the independent
errors algebraic solution for each of the 4 trios possible. Thus, each
classifier will get three estimates for each of the labels.

The alert reader will note the practical value for schemes like AutoML in having a
monitoring algorithm like this. It can be used to error correct bad classifiers.
Independence of errors, not accuracy, is enough to create reliable ensemble
algorithms once you have solved the ground truth inference problem for a task.

The Penn ML Benchmarks use "0" and "1" for the two binary labels. To maintain
consistency with our notation and avoid confusing the labels with the classifier
indices, we use $\alpha$ for the "0" label and $\beta$ for the "1" label.

\subsection{{\tt twonorm} experiment}

The {\tt twonorm} binary classification benchmark consists of 7,400 items (3703/3697) with 20
features. We divided the features randomly into 4 disjoint sets with 5 features each.
Four classifiers were trained, using ``off-the-shelf'' algorithms provided by the {\tt Mathematica}
system: NeuralNetwork, GradientBoostedTrees, NaiveBayes,
and LogisticRegression. Each classifier was trained on 1500 items randomly
selected from a training set of 2000 (1000/1000). We then assembled their decisions
on the remaining items in the benchmark and used these as sole input into
the repeated application of Theorem 1.

\begin{figure}[ht]
\centering
\includegraphics[width=0.45\textwidth]{"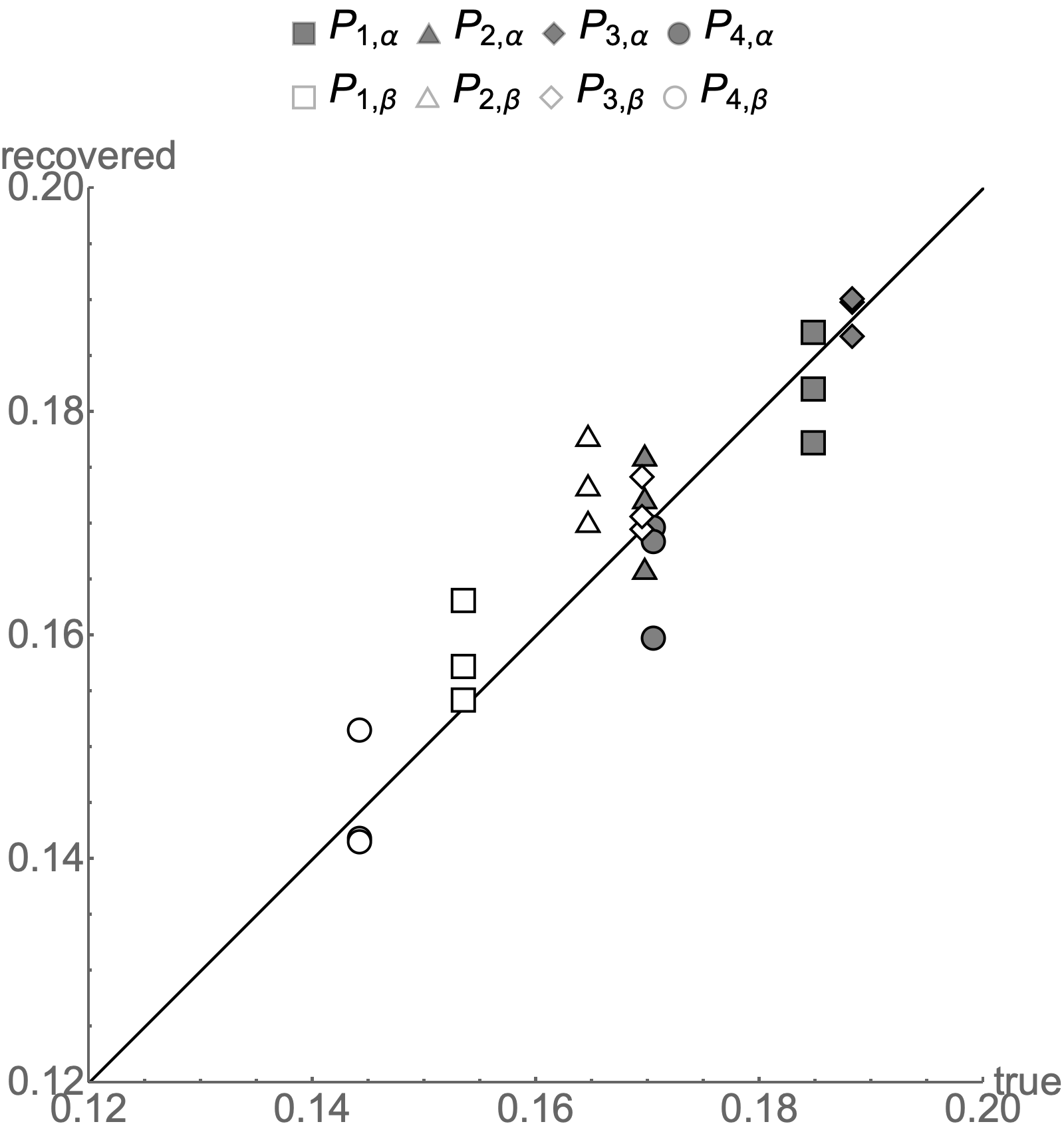"}
\caption{Recovered classifier label accuracy for four classifiers on a single {\tt twonorm} 
experiment}
\label{fig:twonorm-first}
\end{figure}

Our exemplar experiment (Figure \ref{fig:twonorm-first}) shows that the consistency 
between the recovered values is about 1 percent. As noted, each classifier gets three
estimates for each of its label accuracies. On the x-axis we plot the ground truth
value and on the y-axis, the recovered value using the independent polynomial system
solution. For ease of reference, we include the
diagonal line so the reader can see when the recovered value is close to the ground
truth one. For this experiment the
average label correlations are shown in Table~\ref{tbl:twonorm-first}. This experiment is an 
example where both labels had small error correlations and both recovered
estimates were close to the ground truth values for the sample.

\begin{table}[ht]
\caption{Mean and standard deviation for 2-way sample error correlations in the {\tt twonorm} experiment} 
\label{tbl:twonorm-first}
\centering
\begin{tabular}{c|c||c|c}
$\overline{\Gamma_{\alpha;i,j}}$ & $\sigma_{\Gamma_{\alpha;i,j}}$ & $\overline{\Gamma_{\beta;i,j}}$ & $\sigma_{\Gamma_{\beta;i,j}}$ \\
\hline
-0.0000048 & 0.0023 & -0.0021 & 0.0024
\end{tabular}
\end{table}

\subsection{{\tt spambase} experiment}

The {\tt spambase} binary classification benchmark consists of 4601 (2788/1813) items with 57
features. We divided the features randomly into 4 sets with 10 features each.
Four classifiers were trained using the {\tt Mathematica}
system algorithms: NeuralNetwork, SupportVectorMachine, DecisionTree
and NaiveBayes. Each classifier was trained on (200/100) items randomly
selected from a training set of (279/181). These results are an examplar of
how we can recover good estimates for one label with small correlations even
in the presence of larger correlations (and inconsistent estimates) for the
other label.

\begin{figure}[ht]
\centering
\includegraphics[width=0.45\textwidth]{"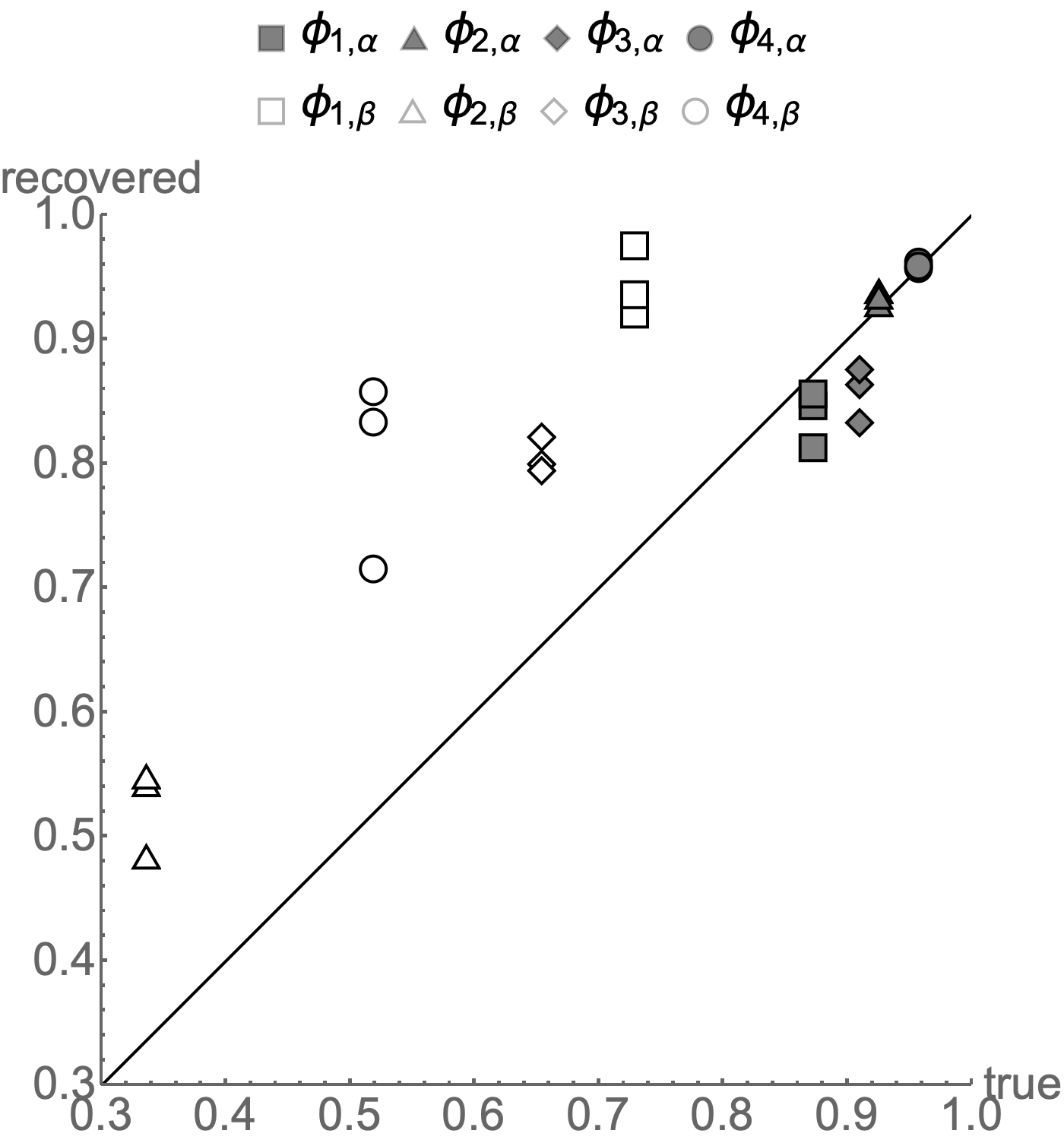"}
\caption{Recovered classifier label accuracy for four classifiers on a single {\tt spambase} 
experiment}
\label{fig:spambase-single}
\end{figure}

\begin{table}[ht]
\caption{Mean and standard deviation for 2-way sample error correlations in the {\tt spambase} experiment} 
\label{tbl:spambase-exemplar}
\begin{center}
\begin{tabular}{c|c||c|c}
$\overline{\Gamma_{\alpha;i,j}}$ & $\sigma_{\Gamma_{\alpha;i,j}}$ & $\overline{\Gamma_{\beta;i,j}}$ & $\sigma_{\Gamma_{\beta;i,j}}$ \\
\hline
0.0056 & 0.0036 & 0.067 & 0.020
\end{tabular}
\end{center}
\end{table}

\subsection{{\tt mushroom} experiment}

The {\tt mushroom} binary classification benchmark consists of 8124 (4208/3916) items with 22
features. We divided the features randomly into 4 sets with (6,6,6,5) features each.
Four classifiers were trained using the {\tt Mathematica}
system algorithms: DecisionTree, NaiveBayes, NeuralNetwork, SupportVectorMachine. 
Each classifier was trained on (100/100) items randomly
selected from a training set of (421/392). These results are an exemplar of roughly
equally noisy recovery for both labels.

\begin{figure}[ht]
\centering
\includegraphics[width=0.45\textwidth]{"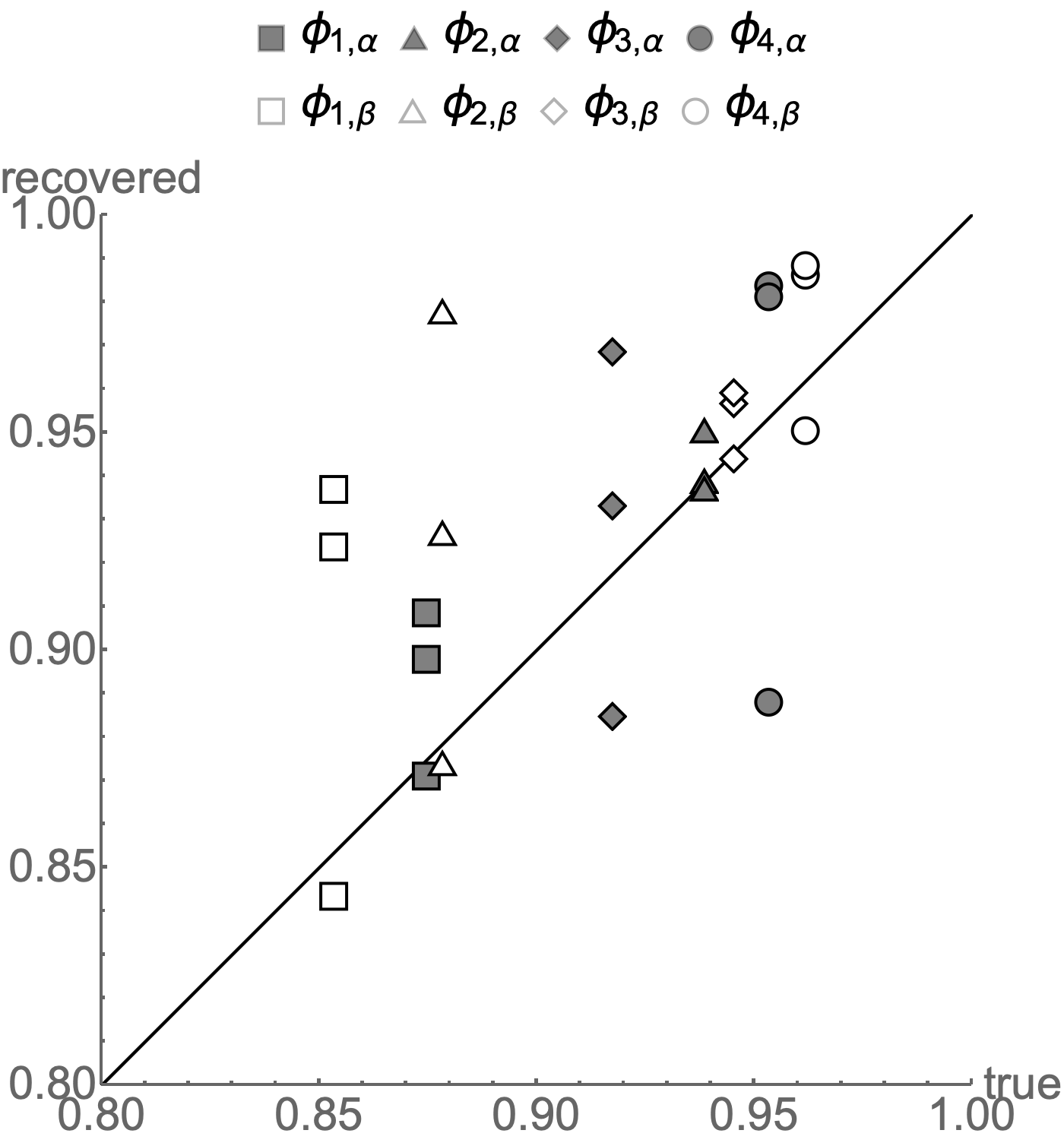"}
\caption{Recovered classifier label accuracy for four classifiers on a single {\tt mushroom} 
experiment}
\label{fig:mushroom-single}
\end{figure}

\begin{table}[ht]
\caption{Mean and standard deviation for 2-way sample error correlations in the {\tt spambase} experiment} 
\label{tbl:mushroom-exemplar}
\centering
\begin{tabular}{c|c||c|c}
$\overline{\Gamma_{\alpha;i,j}}$ & $\sigma_{\Gamma_{\alpha;i,j}}$ & $\overline{\Gamma_{\beta;i,j}}$ & $\sigma_{\Gamma_{\beta;i,j}}$ \\
\hline
0.012 & 0.011 & 0.017 & 0.025
\end{tabular}
\end{table}

\section{The four consistent trios theorem for regressors}

The four consistent trios conjecture remains unsolved for binary
classifiers. We now discuss and
resolve the conjecture for noisy regressors. The ground truth inference
problem we will be solving consists of recovering the precision error
covariance matrix for scalar regressors. This was solved by Corrada and
Schults (2008) in the context of fusing multiple Digital Elevation Models
(DEMs). Here we are interested in the task for theoretical purposes.
If we did not have such a sparsely-correlated recovery method, could we
use something like the four consistent trios to convince ourselves that
the regressors were nearly independent on their sample errors?

\subsection{A ground truth invariant equation for a pair of regressors}

For ease of illustration we will be discussing scalar regressors, but
the same methodology can be applied to the multi-dimensional case. The
ground truth for this task is the set $\{ y_m \}_{m=1}^{M}.$ We also have the set 
$\{y_{m,i}\}_{m=1}^{M}$ for any member of the ensemble of noisy regressors. 
The ground truth inference problem we are trying to solve is the recovery of the
error covariances defined by,
\begin{equation}
    \epsilon_{i,j} = \frac{1}{M} \sum_{m=1}^M \left(y_m - y_{m,i} \right) \left(y_m - y_{m,j} \right).
\end{equation}
The following theorem is easily proved and forms the basis for constructing a
ground-truth invariant linear system of equations that can be used to recover
the error covariances.
\begin{theorem}
Given the estimates $\{y_{m,i}\}$ and $\{y_{m,j}\}$ by two regressors i and j, the
following equation holds,
\begin{equation}
    \epsilon_{i,i} + \epsilon_{j,j} - 2 \, \epsilon_{i,j} = \\
    \frac{1}{M} \sum_{m=1}^M y_{m,i}^2 + \frac{1}{M} \sum_{m=1}^M y_{m,j}^2 - 2 \,  \frac{1}{M} \sum_{m=1}^M y_{m,i} y_{m,j}
    \label{eq:pair-regressors}
\end{equation}
\end{theorem}
The rhs of equation \ref{eq:pair-regressors} is independent of any knowledge of
the ground truth as it only involves moments of the estimates given by the two
regressors. Furthermore, it is invariant to arbitrary global shifts for each
regressor, $\{y_{m,i}\} \longrightarrow \{y_{m,i} + a\}.$ 

\subsection{The under-determined linear system for solving the regressors
GTI problem}

Equations of the form in Equation \ref{eq:pair-regressors} give us
$n(n-1)/2$ equations for the $n(n+1)/2$ terms needed to recover the
precision error covariance matrix.
Why do we say \emph{precision error} and not just \emph{error}?
The constructed linear system is invariant to arbitrary global shifts.
So its use cannot recover the full error covariance matrix but is, instead, 
recovering the precision error covariance matrix. This is equivalent to 
saying that we are computing with de-meaned signals for all of the regressors.

\subsection{The GTI solution for three independent regressors}

For regressors, we have the same phenomena as for binary classifiers.
At some large enough ensemble , an independent
errors model becomes solvable.

Consider independent regressors, at $n=3$
the number of unknowns is equal to the number
of pairs. We just have to solve the system,
\begin{equation}
    \begin{pmatrix}
    1 &  1 & 0 \\
    1 & 0 & 1\\
    0 & 1 & 1
    \end{pmatrix}
    \begin{pmatrix}
    \epsilon_{1,1}\\ \epsilon_{2,2} \\ \epsilon_{3,3}
    \end{pmatrix} =
    \begin{pmatrix}
      \Sigma_{1,2} \\ \Sigma_{1,3} \\ \Sigma_{2,3}
    \end{pmatrix}
    \label{eq:ind-three-regressors}
\end{equation}

\subsection{All trios consistent for four scalar regressors implies homogeneous
error correlations or vanishing data moments}

Now suppose that we are given 4 regressors and we pretend they are independent in
their sample errors so we go ahead and solve four times a system of the form given in
Equation \ref{eq:ind-three-regressors}. The following theorem
is easy to prove as a linear system enforcing the consistency between the four
independent model solutions,
\begin{theorem}
   Four trio solutions for the independent GTI system \ref{eq:ind-three-regressors}
   are consistent only if,
   \begin{equation}
   \epsilon_{i,j} + \epsilon_{k,l} = \epsilon_{i,k} + \epsilon_{j,l}.
    \label{eq:cross-terms}
    \end{equation}
    \label{th:homogeneous}
\end{theorem}

The above theorem does not resolve the conjecture for regressors but it makes
it plausible. Certainly homogeneous solutions, including zero, satisfy this
condition. But can non-homogeneous cross-correlations satisfy this
equation? The answer is yes for consistent trios when 4 or more regressors
are involved.
\begin{theorem}
    Aside from homogeneously constant cross-correlations, scalar regressors
    are consistent if all possible 4 distinct classifiers $\{i,j,k,l\}$ for
    $n>=4$ classifiers satisfy,
    \begin{equation}
        \sum_{d=1}^D (y_{d,i} - y_{d,k})(y_{d,j} - y_{d,l}) = 0
    \end{equation}
    \label{th:data-moments}
\end{theorem}
The immediate consequence of this is that we can quickly check, using
data alone, if Theorem \ref{th:data-moments} applies. If not, it requires further
theoretical investigation to ascertain when consistent solutions are more
likely to imply zero or small correlations than some large homogeneous value.

\section{Conclusions}
We have shown an algebraic way of solving the problem of self-assessing three independent binary classifiers
quickly. Due to its algebraic nature, it can detect non-independent classifiers. And by using it on all
four trios of an ensemble of four, we showed how it could detect when it was working on three Penn ML
classification benchmarks. A similar consistency conjecture for scalar regressors was shown to require
perfectly homogeneous cross-correlations, an unlikely balance point that requires further
theoretical elucidation. For both classifiers and regressors we were able to build tests that can confirm
the independence of the algorithms without the ground truth for the correct answers.

We intend to continue this work by exploring further the prevalence quadratic to see how it can be used
to develop functional forms that measure the error correlations.

\bibliography{IndependenceTestsGTI}

\appendix
\section{Proofs}

\subsection{Proof of Theorem 1}

\textit{Theorem 1, Whenever three independent binary classifiers have 
\decisioneventcounts not all equal to each other and all non-zero, they satisfy a solvable quartic
polynomial system with two solutions, one of them, with the ground truth values}.

The proof has two parts. The first part is purely algebraic - the proof that
the system of eight quartic polynomials has only two point solutions. This is
done with Buchberger's algorithm and inspection of the Gr\"obner Basis we get
as its output. The second part consists of showing that the ground truth values 
for the statistics of the sample we seek to estimate, the prevalence, and classifier
accuracies of the ensemble of three classifiers, is one of these two solutions.

\subsubsection{The quartic system has two point solutions}
The condition that not all \decisioneventcounts are  equal prevents the degenerate
case of all classifiers equally good on both labels. This is a clearly unrealistic
finite sample - any prevalence would satisfy it. Similarly, loss of information
due to few counts would also lead to degenerate cases that are easy to detect. For
example, they could just agree on everything. Those are of mathematical, not
practical interest.
We exclude them so we can talk about the case of practical interest - not all
counts are equal.

This proof is by construction, an application of Buchberger's algorithm for computing the
Gr\"obner Basis to the quartic system. This can be done, for example, with the command {\tt GrobnerBasis}
in the {\tt Mathematica} software system. 
we obtain a quadratic equation in the unknown prevalence \prva by using the following order of the
variables,
\begin{equation}
    \left\{\phi _{3,\beta },\phi _{3,\alpha },\phi _{2,\beta },\phi _{2,\alpha },\phi _{1,\beta },\phi _{1,\alpha },\phi _{\alpha }\right\}
\end{equation}
and a lexicographic order. This single quadratic is the start of two distinct chains of
polynomials; one for the $\alpha$ label, the other for $\beta.$
The $\alpha$ chain allows us to sequentially solve for the \setalphaaccuracies{3}, the other solves
for the \setbetaaccuracies{3}.
All the polynomials in these two chains are linear equations
in the unknown variable one being solved for. The variables involved are detailed in Table \ref{tab:elimination-chain}.
The conclusion is immediate.
There are only two point solutions - each starting with one of the two
values that solve the prevalence quadratic formula.

\begin{table}[ht]
    \centering
    \begin{tabular}{|c|c|}
    \hline
    $\alpha$ &  $\beta$\\
    \hline
    \multicolumn{2}{|c|}{\prva}\\
    \hline
    \prva, $\phi_{3,\alpha}$ & \prva, $\phi_{3,\beta}$ \\
    \hline
    $\phi_{3,\alpha}$, $\phi_{2,\alpha}$ &$\phi_{3,\beta}$, $\phi_{2,\beta}$ \\
    \hline
    $\phi_{2,\alpha}$, $\phi_{1,\alpha}$ &$\phi_{2,\beta}$, $\phi_{1,\beta}$ \\
    \hline
    \end{tabular}
    \caption{Variables in the two elimination chains}
    \label{tab:elimination-chain}
\end{table}

\subsubsection{For independent classifiers, the ground truth values are one
of those two solutions to the prevalence quadratic}
If the classifiers are independent in their sample errors, then the observed
counts of any decision event can be factored in the following form,
\begin{equation}
    \frac{n(\ell_1, \ell_2, \ell_3 \, | \, \labeltrue)}{\nlabeltrue}
    = \frac{n(\ell_1 \, | \, \labeltrue)}{\nlabeltrue}\frac{n(\ell_2 \, | \, \labeltrue)}{\nlabeltrue}
    \frac{n(\ell_3 \, | \, \labeltrue)}{\nlabeltrue}
    \label{eq:independent-factorization}
\end{equation}
But these can all be expressed as the $\phi_{i,\ell}$ terms in the quartic polynomials. 
Since we are assuming the binary labels are disjoint for any item in the sample, every
decision count observed in the sample is equal to
\begin{equation}
    m_{\ell_1, \ell_2, \ell_3} = n(\alpha)\phi_{1,\ell_1,\alpha}\phi_{2,\ell_2,\alpha}\phi_{3,\ell_3,\alpha}
    + n(\beta)\phi_{1,\ell_1,\beta}\phi_{2,\ell_2,\beta}\phi_{3,\ell_3,\beta}.
\end{equation}
Dividing by $M$ and rewriting all the $\phi_{i,\ell_1,\ell_2}$ terms of the classifier accuracies,
\begin{align}
    \phi_{i,\beta,\alpha} = & 1 - \phi_{i,\alpha}\\
    \phi_{i,\alpha,\beta} = & 1 - \phi_{i,\beta}
\end{align}
yields the quartic system of Theorem 1. Hence the ground truth values solve the system and it must be one
of the two solutions to the polynomial system.

\subsubsection{What this theorem is not saying}
The assumption stated in Equation \ref{eq:independent-factorization} gives the whole game away.
If we are allowed to assume this, it removes any finite sample, however it was produced,
that does not meet this factorization assumption. The interesting part in practical terms
is that, for small enough samples, this will be hard to do if, say, we start with an independent
generative process.
Nonetheless, independence theorems like this are always interesting because they can start
as a foundation for a sparsely or weakly correlated approach. For this more practical
task, one needs Theorem 2.

One can get a feel for how easy it is to create a small sample for it by considering random
accuracies of either $1/3$ or $2/3$ and pretend the prevalence is $1/2$. On one simulation
of this procedure using Theorem 1 to produce the decision event counts, $M=18$ was enough
to create a finite sample that obeys the Theorem. But most of the $8^{18} = 18,014,398,509,481,984$ 
ways three binary classifiers could have labeled the sample would not satisfy Theorem 1.

Hence our interest in handling non-zero correlations. They can occur due to small sample
size or because asymptotically the classifiers are correlated. In either case,
a useful non-parametric method has to handle non-zero error correlations.

\subsection{Proof of Theorem 2}

\textit{Theorem 2, A complete polynomial description of the $L^n$ decision event counts for $n$
arbitrarily correlated L-label classifiers is possible. The system is of polynomial order $n+1$ 
and the ground truth values are in the algebraic variety of this complete description.}

By the disjointness of labels, any of the $L^n$ decision even counts can be expressed as,
\begin{equation}
    m_{\ell_1, \ell_2, \ldots, \ell_n} = \sum_{l=1}^L m(\ell_1, \ell_2, \ldots, \ell_n \, | \, \ell_l).
\end{equation}
This can be rewritten as,
\begin{equation}
    m_{\ell_1, \ell_2, \ldots, \ell_n} = \sum_{l=1}^L \frac{m(\ell_1, \ell_2, \ldots, \ell_n \, | \, \ell_l)}{m(\ell_l)} m(\ell_l).
\end{equation}
The rest of the proof can now proceed by focusing solely on how to write any $m(\ell_1, \ell_2, \ldots, \ell_n \, | \, \ell_l)$
in terms of the prevalences, marginal accuracies, and the error moments for that label alone.

\subsubsection{Error moments for arbitrary number of labels}
For arbitrary number of labels it makes more sense to talk about error moments instead of correlations as in the
binary case ($L=2$).
These error moments are numerous since we must take into account $L^n$ decision
event counts for a given true label. The general definition of the error moments is,
\begin{equation}
    \Gamma_{\ell_i, \ell_j, \ldots; \elltrue} = \frac{1}{m(\ell_{\text{true}})} \sum_{d=1}^{m(\elltrue)} 
    (x_{d,\ell_i,\elltrue} - \phi_{\ell_i,\elltrue})(x_{d,\ell_j,\elltrue} - \phi_{\ell_j,\elltrue})
\end{equation}
Note that any decision even count for a label can be written as the sum, over the sample items, of evaluating a
monomial of the form,
\begin{equation}
    x_{d,\ell_1, \elltrue} \ldots x_{d,\ell_n, \elltrue},
    \label{eq:decision-momonial}
\end{equation}
where the $x_{d,\ell_i, \elltrue}$ are indicator variables for making the $\ell_i$ decision on an item of the true label,
\elltrue.
These per label indicator variables for each classifier satisfy,
\begin{equation}
    \sum_{\ell_i} x_{d,\ell_i,\elltrue} = 1.
\end{equation}

This condition and the normalization condition for marginal accuracies of each classifier,
\begin{equation}
    \sum_{\ell_i} \phi_{\ell_i,\elltrue} = 1
\end{equation}
means that these error moments are related. For example, in the binary label case a pair of classifiers has four
error moments for each true label. But they can all be reduced to a single one as shown below for the case of
the $\alpha$ label
\begin{align}
    \Gamma_{\alpha, \beta; \alpha} = & - \Gamma_{\alpha, \alpha; \alpha}\\
    \Gamma_{\beta, \alpha; \alpha} = & - \Gamma_{\alpha, \alpha; \alpha}\\
    \Gamma_{\beta, \beta; \alpha} = &  \Gamma_{\alpha, \alpha; \alpha}
\end{align}

\subsubsection{All decision event counts can be written in terms of the error moments and
marginal accuracies}
The crucial point is that all decision event counts for a true label can be rewritten in terms of
these error moments and the marginal accuracies of the classifiers. Starting with the full monomial
shown in Equation \ref{eq:decision-momonial}, we can rewrite a count in terms of the n-way correlation
moment and lower order monomials of the indicator functions. But these lower order monomials can be, in turn, rewritten
in terms of lower order correlations and so on. This descent ends at the marginal accuracies of the
classifiers.

The rest is algebra. 

By construction, the ground truth values for all these unknown sample statistics
will satisfy these polynomial equations exactly. Therefore, the ground truth values must reside in
the set of points that satisfy the polynomial system. This is called the algebraic variety of a
polynomial system. QED.

\subsubsection{The binary label case}

To illustrate the above generic observation we will illustrate how the computation proceeds for the
binary label case. To simplify the heavy notation that is required in the general case we use,
\begin{equation}
    x_{d,\ell_i = \elltrue, \elltrue} = x_{d,i},
\end{equation}
when discussing binary classifiers.

Any decision event count for a label in binary classification can be written as products of $x_{d,i}$ 
and $(1-x_{d,i})$ depending on whether a classifier has the label correct or not. So any of these 
label ensemble decision counts can be written as monomials involving, at most, n powers of the 
$x_{d,i}$. We can use all the error correlations to rewrite any of
these monomials. We sketch how this can be done. To simplify the notation further, we
will omit the sum over the sample items. The presence of the subscript $d$ is to be taken as implying 
that omitted sum.

Consider the case of two classifiers, $n=2$, and the $m(\alpha, \beta \, | \, \alpha)$ decision event 
count. This is given by,
\begin{equation}
    x_{d,1} (1 - x_{d,2}) = m_{\alpha} \phi_{1,\alpha} -  x_{d,1}  x_{d,2}.
\end{equation}
But the sample sum of $x_{d,i}  x_{d,j}$ can be re-written using the 2-way correlation term,
\begin{equation}
    m_{\alpha} \Gamma_{i,j;\alpha} = (x_{d,i} - \phi_{i,\alpha})(x_{d,j} - \phi_{j,\alpha})
\end{equation}
to give us,
\begin{equation}
    x_{d,i} x_{d,j} = m_{\alpha} \left(\Gamma_{i,j;\alpha} - \phi_{i,\alpha} \phi_{j,\alpha} \right)
\end{equation}
Putting together these equations we can write the label ensemble decision count for the $\alpha$
items in the sample as,
\begin{equation}
    m(\alpha, \beta \, | \, \alpha) = m_{\alpha}\left( \phi_{1,\alpha} (1-\phi_{2,\alpha}) - \Gamma_{1,2;\alpha} \right).
\end{equation}
We can immediately generalize this to obtain the exact polynomial descrition of the $m(\alpha, \beta)$ 
count for any two arbitrarily correlated classifiers,
\begin{equation}
    m(\alpha, \beta) = \left( \phi_{1,\alpha} (1-\phi_{2,\alpha}) - \Gamma_{1,2;\alpha} \right) m(\alpha) + 
     \left( (1 - \phi_{1,\beta}) \phi_{2,\beta} - \Gamma_{1,2;\beta} \right) m(\beta).
\end{equation}

Proceeding to any number of classifiers is just the repeated application of this process.
For illustration purposes, we give here a single one of the eight polynomials that comprise the three 
arbitrarily correlated binary classifiers,
\begin{multline}
    \phi _{\alpha } \left(\left(1-\phi _{1,\alpha }\right) \phi
   _{2,\alpha } \left(1-\phi _{3,\alpha }\right) -\Gamma _{1,2,\alpha } \left(1-\phi _{3,\alpha
   }\right)+\Gamma _{1,3,\alpha } \phi _{2,\alpha }-\Gamma _{2,3,\alpha } \left(1-\phi _{1,\alpha }\right)+\Gamma _{1,2,3,\alpha }\right) + \\
   \left(1-\phi _{\alpha }\right) \left(\phi _{1,\beta } \left(1-\phi _{2,\beta }\right) \phi _{3,\beta } -\Gamma _{1,2,\beta } \phi _{3,\beta }+\Gamma _{1,3,\beta } \left(1-\phi _{2,\beta }\right)-\Gamma _{2,3,\beta } \phi _{1,\beta
   }-\Gamma _{1,2,3,\beta }\right)
\end{multline}
We leave it as an exercise for the reader to guess which ensemble decision event count is described by this 
polynomial.

\subsection{Proof of Theorem 3, correlated classifiers have an irreducible
square root in the independent solution for the prevalence}

The Gr\"obner basis for the independent classifiers polynomial system
contains a quadratic that solves for the unknown \prva, as detailed in the proof of Theorem 1.
It can be represented as follows,
\begin{equation}
    \acoeff \prva^2 + \bcoeff \prva + \ccoeff = 0,
\end{equation}
The \acoeff, \bcoeff, and \ccoeff polynomials are quite large so it is not illustrative to write them out.
Table \ref{tab:coeff-properties} shows some of their properties. \ccoeff is the only term that is 
factorizable when using the $m$ variables.
\begin{table}[ht]
    \centering
    \begin{tabular}{c|c|c}
       coefficient & number of terms &  irreducible? \\
       \hline
        \acoeff & 72 & yes \\
        \hline
        \bcoeff & 72 & yes \\
        \hline
        \ccoeff & 17/17/17 & no
    \end{tabular}
    \caption{Some algebraic properties of the polynomial coefficients of the prevalence quadratic}
    \label{tab:coeff-properties}
\end{table}

The proof then rests on the irreducibility of the square root term in the quadratic formula,
\begin{equation}
    \sqrt{b^2 - 4 a c}.
\end{equation}
If the classifiers are independent on the sample, then the independent polynomial system is an
exact expression of the observed ensemble decision event counts. This allows us to re-express
the counts as polynomials of the \prva, \setalphaaccuracies{3}, and \setbetaaccuracies{3}.
The resulting expression for the quadratic square root is then a perfect square as described
in the paper. The factor is precisely the one needed to obtain either \prva or $1-\prva$ as
the solution depending on whether we use the plus or minus sign for the square root term.

But Theorem 2 gives us a set of eight polynomials that allow us to express the observed counts
for any amount of correlation between the classifiers, not just the independent case.
This transforms the square root term in the independent solution into a polynomial expression 
in the variables \prva, \setalphaaccuracies{3}, \setbetaaccuracies{3}, \varscorrtwo, and \varscorrthree.
As described in the paper, the resulting factorization of $b^2-4 a c$ leads to this
very general form under the square root operation,
\begin{equation}
    p_{1}(\prva, \setalphaaccuracies{3}, \setbetaaccuracies{3}, \varscorrtwo, \varscorrthree)^2
    p_{2}(\prva, \setalphaaccuracies{3}, \setbetaaccuracies{3}, \varscorrtwo, \varscorrthree),
\end{equation}
where $p_1$ and $p_2$ are irreducible polynomials in their arguments. $p_2$ is an irreducible
polynomial with 72 terms, monomials of the \decisioneventcounts. Substitution of
the full polynomial system for three classifiers transform this polynomial into an irreducible
polynomial with 2,411 terms (it can be written as the sum of factorizable factors to make it
more manageable). Since the irreducibility is due to the correlation terms, one can explore
what transformations would make it reducible by making simplifying assumptions. We show some
of these in Table \ref{tab:irreducibility-transformations}. Some of the assumptions led to
further factorization but still retained a large irreducible polynomial component. We must
conclude that obtaining an integer ratio solution for correlated classifiers is rare and further
theorems are needed to quantify this unlikely event. 
All of our experiments with the Penn ML benchmarks contained an irreducible square root in
the independent solution to the prevalence.
\begin{table}[ht]
    \centering
    \begin{tabular}{c|c|c}
       transformation  &  irreducible? & length\\
       \hline
       all label pair correlations equal  & yes & $1830$\\
       all pair correlations equal & yes & $1206$\\
       all 2-way and 3-way correlations equal & yes & $1122$\\
       one label correlations zero & no & $2/516$ \\
       one label zero, other constant & no & $2/436$\\
       all correlations equal & yes & $1039$\\
       all correlations and accuracies equal & yes & $48$
    \end{tabular}
    \caption{Simplifying assumptions for unknown correlations and their effect on the irreducibility of
    the square root in the quadratic formula for the independent prevalence solution.}
    \label{tab:irreducibility-transformations}
\end{table}

\subsubsection{Addendum: the square root can be imaginary or lead to unphysical values}
We did not discuss in the paper, because of space considerations,
another mathematical possibility for the independent system solution.
It could result in complex solutions or values that lie outside the (0,1) range. None
of these solutions make any sense as values for the accuracies or prevalence since
these must all lie within the (0,1) interval. Both of these cases were regularly encountered
during the Penn ML Benchmark experiments whenever the classifiers were too correlated in their
errors.

\subsubsection{Addendum: Taylor expanding the independent solution about the unknown error correlations}
Future work along this algebraic approach will focus on the obvious application of Theorem 1 and Theorem 2 -
Taylor expand the independent solution about zero error correlations so that we can use the inconsistencies
between independent polynomial solutions as a way of estimating the correlations themselves.

\subsection{Proof of Theorem 3}

\textit{Theorem 3, There exists a linear equation for a pair of scalar regressors that is
independent of the ground truth values}.

Consider two scalar regressors. We can write their error correlations as
\begin{align}
    \epsilon_{i,i} = & \sum_{d=1}^D (y_d - y_{d,i})^2 \\
    \epsilon_{j,j} = & \sum_{d=1}^D (y_d - y_{d,j})^2 \\
    \epsilon_{i,j} = & \sum_{d=1}^D (y_d - y_{d,i})(y_d - y_{d,j}).
\end{align}
Abusing notation, we can see that these three equations contain three terms
involving the unknown ground truth: $y_{d}^2$, $y_{d}y_{d,i}$, and $y_{d}y_{d,j}.$
We can eliminate these three unknowns using the three equations above to obtain
the required linear equation that is independent of ground truth,
\begin{equation}
    \epsilon_{i,i} + \epsilon_{j,j} - 2 \epsilon_{i,j} = \sum_{d=1}^D y_{d,i}^2 + \sum_{d=1}^D y_{d,j}^2 -
    \sum_{d=1}^D y_{d,i} y_{d,j}
\end{equation}

\subsection{Proof of Theorem 4, purely algebraic part}

\textit{Theorem 4, Consistency of independent solutions for four regressors implies,
\begin{equation}
    \epsilon_{i,j} + \epsilon_{k,l} = \epsilon_{i,k} + \epsilon_{j,l},
\end{equation}
for all permutations of the four indices $\{i,j,k,l\}$}

Consistency of four trios is equivalent to the following linear system of six equations for
the six possible pairs for four regressors,
\begin{align}
    \epsilon_{i,i} + \epsilon_{j,j} - 2 \epsilon_{i,j} & =  \epsilon_{i,i}^{(\text{ind})} + \epsilon_{j,j}^{(\text{ind})}\\
    \epsilon_{i,i} + \epsilon_{k,k} - 2 \epsilon_{i,k} & =  \epsilon_{i,i}^{(\text{ind})} + \epsilon_{k,k}^{(\text{ind})}\\
    \epsilon_{i,l} + \epsilon_{l,l} - 2 \epsilon_{i,l} & =  \epsilon_{i,i}^{(\text{ind})} + \epsilon_{l,l}^{(\text{ind})}\\
    \epsilon_{j,j} + \epsilon_{k,k} - 2 \epsilon_{j,k} & =  \epsilon_{j,j}^{(\text{ind})} + \epsilon_{k,k}^{(\text{ind})}\\
    \epsilon_{j,j} + \epsilon_{l,l} - 2 \epsilon_{j,l} & =  \epsilon_{j,j}^{(\text{ind})} + \epsilon_{l,l}^{(\text{ind})}\\
    \epsilon_{k,k} + \epsilon_{l,l} - 2 \epsilon_{k,l} & =  \epsilon_{k,k}^{(\text{ind})} + \epsilon_{l,l}^{(\text{ind})},
\end{align}
where the $\epsilon_{i,i}^{(\text{ind})}$ are the consistent independent system solutions. Straightforward linear
algebra leads to the elimination of all of the $\epsilon_{i,i}^{(\text{ind})}$ as well as the $\epsilon_{i,i}.$
This yields the consistency constraint equations,
\begin{equation}
    \epsilon_{i,j} + \epsilon_{k,l} = \epsilon_{i,k} + \epsilon_{j,l} = \epsilon_{i,l} + \epsilon_{j,k}
\end{equation}
\subsection{Proof of Theorem 5}

\textit{Theorem 5, The solutions to Theorem 4 are either all cross-correlations are equal or
all the following data moments are zero
\begin{equation}
    \sum_{d=1}^D (y_{d,j} - y_{d,k})  (y_{d,i} - y_{d,l}) = 0
\end{equation}}

As shown in Theorem 4, for all $\{i,j,k,l\}$ the consistency of independent solutions requires
\begin{equation}
    \epsilon_{i,j} + \epsilon_{k,l} = \epsilon_{i,k} + \epsilon_{j,l}.
\end{equation}
You can turn the above equation into a statement of data moments
alone by using the definition of cross-correlation. All the
terms in the cross-correlations involving the unknown ground truth
cancel. The final result can be expressed solely in terms of the
estimates by the four regressors as,
\begin{equation}
    \sum_{d}(y_{d,j}-y_{d,k})(y_{d,i}-y_{d,l}) =  0
\end{equation}

Therefore we can check, without knowledge of the ground truth, if all
these relations hold for any permutation of the $i$, $j$, $k$, and $l.$
This is clearly a very particular condition for situations other than
constant cross-correlations. Therefore, we must conclude that we are
left with a situation similar to other theorems in the paper - consistency
is possible with non-zero correlations but highly unlikely for most
practical cases. This conclusion must be clarified by further theoretical
analysis.

As we have pointed out in the paper, the case of sparsely correlated regressors
is solvable by using Compressed Sensing algorithms so the solution for practical
cases near independence already exists.

\subsubsection{The conjecture for binary classifiers}
This approach can also be followed for the conjecture as it applies to binary classifiers.
In that case we have 4 sets of eight polynomial equations each. On the LHS are the full
polynomial systems of Theorem 2. On the RHS are the solutions obtained by pretending
that the classifiers are independent. We have been unable to resolve the consequences
of this system of 32 quartic equations using a brute-force application of Buchberger's
algorithm.

\section{General comment about ground truth statistics yet to be described}
It is clear by this algebraic formulation that we have shown how to derive
a particular set of ground truth statistics. The two GTI problems discussed
here are just exemplar of a similar algebraic approach that can be applied to other ground
truth statistics. We mention one only - sequential error statistics. In
applications such as DNA sequences, the accuracy of pairs or triplets of
bases may be an important consideration for some task requiring high precision.

\section{Details of the Penn ML Benchmarks experiments}

Here we provide more detail about the Penn ML Benchmarks experiments
and include the ground truth counts for each of the experiments. These counts
are a sufficient statistic for all relevant quantities needed for an experiment.
From these tables one can calculate any of the ground truth statistics we discussed
in this paper - the prevalence, the label accuracies of the classifiers, and
their sample error correlations up 2 to 4-way.

All experiments used the {\tt Mathematica} v12.1.1.0 software package, a natural choice for
us given the combination of algebraic and classification computations needed to
carry out the theoretical and experimental work we have presented.
Unless otherwise stated, we used default versions for the classifiers provided by
{\tt Mathematica}.

\subsection{The {\tt two-norm} experiment}

The classifiers used were,
\begin{enumerate}
    \item {\tt NeuralNetwork}, with $\text{NetworkDepth} = 5$
    \item {\tt GradientBoostedTrees}
    \item {\tt NaiveBayes}
    \item {\tt LogisticRegression}
\end{enumerate}
The ground truth counts are shown in Table \ref{tab:twonorm-counts}.

\begin{table}[ht]
    \centering
    \begin{tabular}{c|c|c}
    decision event & 0 label & 1 label \\
    \hline
 \{0,0,0,0\} & 4 & 1330 \\
   \hline
 \{0,0,0,1\} & 7 & 237 \\
   \hline
 \{0,0,1,0\} & 18 & 271 \\
   \hline
 \{0,0,1,1\} & 58 & 48 \\
   \hline
 \{0,1,0,0\} & 15 & 285 \\
   \hline
 \{0,1,0,1\} & 58 & 42 \\
   \hline
 \{0,1,1,0\} & 50 & 58 \\
   \hline
 \{0,1,1,1\} & 268 & 7 \\
   \hline
 \{1,0,0,0\} & 11 & 258 \\
   \hline
 \{1,0,0,1\} & 52 & 38 \\
   \hline
 \{1,0,1,0\} & 42 & 58 \\
   \hline
 \{1,0,1,1\} & 247 & 8 \\
   \hline
 \{1,1,0,0\} & 56 & 38 \\
   \hline
 \{1,1,0,1\} & 284 & 7 \\
   \hline
 \{1,1,1,0\} & 245 & 5 \\
   \hline
 \{1,1,1,1\} & 1172 & 1 \\
\end{tabular}
    \caption{Observed decision event counts by true label for the ensemble of four classifiers in the {\tt twonorm} exemplar experiment}
    \label{tab:twonorm-counts}
\end{table}

\subsection{The {\tt spambase} experiment}

The classifiers used were,
\begin{enumerate}
    \item {\tt NeuralNetwork}, with $\text{NetworkDepth} = 4$
    \item {\tt SupportVectorMachine}, with $\text{KernelType} = \text{Polynomial}$, and $\text{PolynomialDegree} = 3$
    \item {\tt DecisionTree}, with $\text{DistributionSmoothing} = 5$
    \item {\tt NaiveBayes}
\end{enumerate}
The ground truth counts are shown in Table \ref{tab:spambase-counts}.
\begin{table}[ht]
    \centering
    \begin{tabular}{c|c|c}
    decision event & 0 label & 1 label \\
    \hline
 \{0,0,0,0\} & 1827 & 185 \\
   \hline
 \{0,0,0,1\} & 53 & 25 \\
   \hline
 \{0,0,1,0\} & 145 & 153 \\
   \hline
 \{0,0,1,1\} & 13 & 30 \\
   \hline
 \{0,1,0,0\} & 121 & 14 \\
   \hline
 \{0,1,0,1\} & 17 & 12 \\
   \hline
 \{0,1,1,0\} & 9 & 14 \\
   \hline
 \{0,1,1,1\} & 3 & 10 \\
   \hline
 \{1,0,0,0\} & 223 & 151 \\
   \hline
 \{1,0,0,1\} & 10 & 90 \\
   \hline
 \{1,0,1,0\} & 45 & 182 \\
   \hline
 \{1,0,1,1\} & 5 & 268 \\
   \hline
 \{1,1,0,0\} & 26 & 22 \\
   \hline
 \{1,1,0,1\} & 5 & 66 \\
   \hline
 \{1,1,1,0\} & 5 & 65 \\
   \hline
 \{1,1,1,1\} & 2 & 345 \\
\end{tabular}
    \caption{Observed decision event counts by true label for the ensemble of four classifiers in the {\tt spambase} exemplar experiment}
    \label{tab:spambase-counts}
\end{table}

\subsection{The {\tt mushroom} experiment}
The classifiers used were,
\begin{enumerate}
    \item {\tt DecisionTree}, with $\text{DistributionSmoothing} = 5$
    \item {\tt NaiveBayes}
    \item {\tt NeuralNetwork}, with $\text{NetworkDepth} = 4$
    \item {\tt SupportVectorMachine}, with $\text{KernelType} = \text{Polynomial}$, and $\text{PolynomialDegree} = 3$
\end{enumerate}
The ground truth counts are shown in Table \ref{tab:mushroom-counts}.
\begin{table}[ht]
    \centering
    \begin{tabular}{c|c|c}
    decision event & 0 label & 1 label \\
    \hline
 \{0,0,0,0\} & 2929 & 0 \\
 \hline
 \{0,0,0,1\} & 75 & 0 \\
 \hline
 \{0,0,1,0\} & 70 & 28 \\
 \hline
 \{0,0,1,1\} & 45 & 266 \\
 \hline
 \{0,1,0,0\} & 135 & 35 \\
 \hline
 \{0,1,0,1\} & 0 & 0 \\
 \hline
 \{0,1,1,0\} & 16 & 14 \\
 \hline
 \{0,1,1,1\} & 42 & 174 \\
 \hline
 \{1,0,0,0\} & 310 & 0 \\
 \hline
 \{1,0,0,1\} & 5 & 0 \\
 \hline
 \{1,0,1,0\} & 110 & 29 \\
 \hline
 \{1,0,1,1\} & 10 & 106 \\
 \hline
 \{1,1,0,0\} & 20 & 29 \\
 \hline
 \{1,1,0,1\} & 0 & 129 \\
 \hline
 \{1,1,1,0\} & 20 & 0 \\
 \hline
 \{1,1,1,1\} & 0 & 2714 \\
\end{tabular}
    \caption{Observed decision event counts by true label for the ensemble of four classifiers in the {\tt mushroom} exemplar experiment}
    \label{tab:mushroom-counts}
\end{table}

\end{document}